\documentclass[10pt,twocolumn,letterpaper]{article}

\usepackage{iccv}
\usepackage{times}
\usepackage{epsfig}
\usepackage{graphicx}
\usepackage{amsmath}
\usepackage{amssymb}

\usepackage{algorithm}
\usepackage{algpseudocode}

\usepackage{booktabs,multirow,rotating}

\def\eg{\emph{e.g}\onedot} 
\def\ie{\emph{i.e}\onedot}

\def\etal{\emph{et al}\onedot}
\makeatother

\def\Vec#1{{\boldsymbol{#1}}}
\def\Mat#1{{\boldsymbol{#1}}}

\usepackage[breaklinks=true,bookmarks=false]{hyperref}

\iccvfinalcopy 

\def\iccvPaperID{797} 

\ificcvfinal\fi

\begin{document}
	
	\title{Graph Convolutional Networks for Temporal Action Localization}
	
	\author{
		Runhao~Zeng$^{1}$\thanks{indicates equal contributions. This work was done when Runhao Zeng was served as a research intern in Tencent AI Lab under the supervision of Wenbing Huang.}~~~~Wenbing~Huang$^{2,5*}$~~~~Mingkui~Tan$^{1,4}$\thanks{Corresponding author}~~~~Yu~Rong$^{2}$\\
		Peilin~Zhao$^{2}$~~~~Junzhou~Huang$^{2}$~~~~ Chuang~Gan$^{3}$\\
		$^{1}$School of Software Engineering, South China University of Technology, China \\ 
		$^{2}$Tencent AI Lab~~~~$^{3}$MIT-IBM Watson AI Lab~~~~$^{4}$Peng Cheng Laboratory, Shenzhen \\
		$^{5}$Department of Computer Science and Technology, Tsinghua University, State Key Lab. of Intelligent \\
		Technology and Systems, Tsinghua National Lab. for Information Science and Technology (TNList) \\
		{\tt\small \{runhaozeng.cs, ganchuang1990\}@gmail.com, hwenbing@126.com,
			mingkuitan@scut.edu.cn}
	}
	
	\maketitle
	\ificcvfinal\thispagestyle{empty}\fi
	
	\begin{abstract}
		Most state-of-the-art action localization systems process each action proposal individually, without explicitly exploiting their relations during learning. However, the relations between proposals actually play an important role in action localization, since a meaningful action always consists of multiple proposals in a video. In this paper, we propose to exploit the proposal-proposal relations using Graph Convolutional Networks (GCNs). First, we construct an action proposal graph, where each proposal is represented as a node and their relations between two proposals as an edge. Here, we use two types of relations, one for capturing the context information for each proposal and the other one for characterizing the correlations between distinct actions. Then we apply the GCNs over the graph to model the relations among different proposals and learn powerful representations for the action classification and localization. Experimental results show that our approach significantly outperforms the state-of-the-art on THUMOS14 (49.1\% versus 42.8\%). Moreover, augmentation experiments on ActivityNet also verify the efficacy of modeling action proposal relationships. Codes are available at \href{https://github.com/Alvin-Zeng/PGCN}{https://github.com/Alvin-Zeng/PGCN}.
		
	\end{abstract}

	\section{Introduction}
	
\begin{figure}[t]
		\centering
		\includegraphics[width=\linewidth]{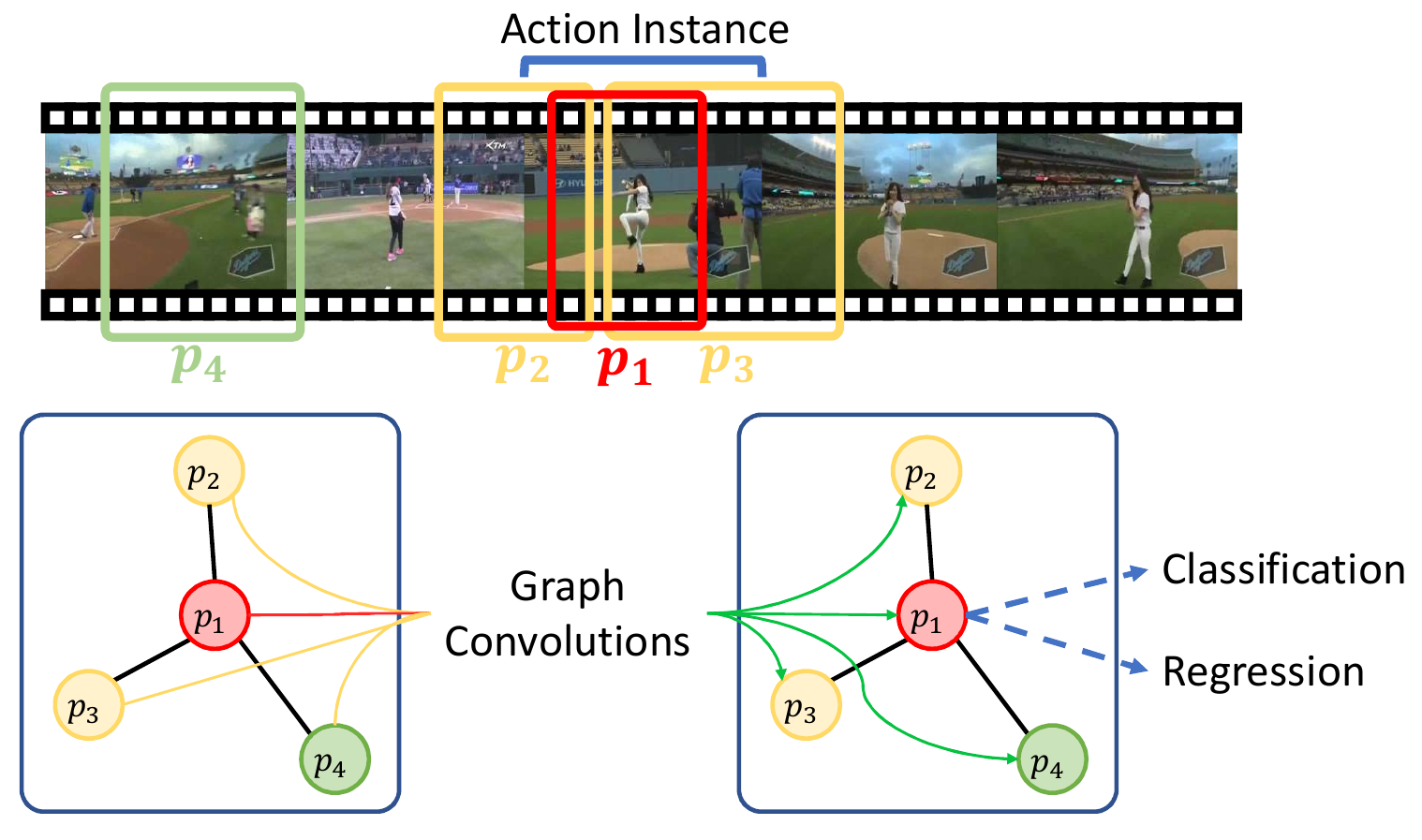}
		\caption{Schematic depiction of our approach. We apply graph convolutional networks to model the proposal-proposal interactions and boost the temporal action localization performance.}
		\label{Fig:simple}
		\vspace{-0.35cm}
	\end{figure}

	Understanding human actions in videos has been becoming a prominent research topic in computer vision, owing to its various applications in security surveillance, human behavior analysis and many other areas~\cite{duan2018weakly,simonyan2014two,tran2015learning,fan2018end,gan2018geometry,gan2015devnet,gan2016recognizing,gan2016you,wang2016temporal}. Despite the fruitful progress in this vein, there are still some challenging tasks demanding further exploration --- \emph{temporal action localization} is such an example. To deal with real videos that are untrimmed and usually contain the background of irrelevant activities, temporal action localization requires the machine to not only classify the actions of interest but also localize the start and end time of every action instance. Consider a sport video as illustrated in Figure~\ref{Fig:simple}, the detector should find out the frames where the action event is happening and identify the category of the event.

	Temporal Action localization has attracted increasing attention in the last several years~\cite{chao2018rethinking,gao2017cascaded,lin2017single,shou2017cdc,shou2016temporal}. Inspired by the success of object detection, most current action detection methods resort to the two-stage pipeline: they first generate a set of 1D temporal proposals and then perform classification and temporal boundary regression on each proposal individually. 
	However, processing each proposal separately in the prediction stage will inevitably neglect the semantic relations between proposals. 
	
	We contend that exploiting the proposal-proposal relations in the video domain provides more cues to facilitate the recognition of each proposal instance. To illustrate this, we revisit the example in Figure~\ref{Fig:simple}, where we have generated four proposals. 
	On the one hand, the proposals $\Mat{p}_1$, $\Mat{p}_2$ and $\Mat{p}_3$ overlapping with each other describe different parts of the same action instance (\ie, the start period, main body and end period).
	Conventional methods perform prediction on $\Mat{p}_1$ by using its feature alone, which we think is insufficient to deliver complete knowledge for the detection.
	If we additionally take the features of $\Mat{p}_2$ and $\Mat{p}_3$ into account, we will obtain more contextual information around $\Mat{p}_1$, which is advantageous especially for the temporal boundary regression of $\Mat{p}_1$.
	On the other hand, $\Mat{p}_4$ describes the background (\ie, the sport field), and its content is also helpful in identifying the action label of $\Mat{p}_1$, since what is happening on the sport field is likely to be sport action (\eg ``discus throwing'') but not the one happens elsewhere  (\eg ``kissing''). In other words, the classification of $\Mat{p}_1$ can be partly guided by the content of $\Mat{p}_4$ even they are temporally disjointed. 
	
	To model the proposal-proposal interactions, one may employ the self-attention mechanism~\cite{vaswani2017attention} --- as what has been conducted previously in language translation~\cite{vaswani2017attention} and object detection~\cite{hu2018relation} --- to capture the pair-wise similarity between proposals. 
	A self-attention module can affect an individual proposal by aggregating information from all other proposals with the automatically learned aggregation weights.
	However, this method is computationally expensive as querying all proposal pairs has a quadratic complexity of the proposal number (note that each video could contain more than thousands of proposals). On the contrary, Graph Convolutional Networks (GCNs) , which generalize convolutions from grid-like data (\eg images) to non-grid structures (\eg social networks), have received increasing interests in the machine learning domain~\cite{kipf2017semi,yan2018spatial}. GCNs can affect each node by aggregating information from the adjacent nodes, and thus are very suitable for leveraging the relations between proposals. More importantly, 
	unlike the self-attention strategy, 
	applying GCNs enables us to aggregate information from only the local neighbourhoods for each proposal, and thus can help decrease the computational complexity remarkably.
	
	In this paper, we regard the proposals as nodes of a specific graph and take advantage of GCNs for modeling the proposal relations. Motivated by the discussions above, we construct the graph by investigating two kinds of edges between proposals, including the \emph{contextual edges} to incorporate the contextual information for each proposal instance (\eg, detecting $\Mat{p}_1$ by accessing $\Mat{p}_2$ and $\Mat{p}_3$ in Figure~\ref{Fig:simple}) and the \emph{surrounding edges} to query knowledge from nearby but distinct proposals (\eg, querying $\Mat{p}_4$ for $\Mat{p}_1$ in Figure~\ref{Fig:simple}). 
	
	We then perform graph convolutions on the constructed graph. Although the information is aggregated from local neighbors in each layer,
	message passing between distant nodes is still possible if the depth of GCNs increases. Besides, we conduct two different GCNs to perform classification and regression separately, which is demonstrated to be effective by our experiments. Moreover, to avoid the overwhelming computation cost, we further devise a sampling strategy to train the GCNs efficiently while still preserving desired detection performance. We evaluate our proposed method on two popular benchmarks for temporal action detection, \ie, THUMOS14~\cite{jiang2014thumos} and AcitivityNet1.3~\cite{caba2015activitynet}.
	
	To sum up, our contributions are as follow:
	\begin{itemize}
		\item To the best of our knowledge, we are the first to exploit the proposal-proposal relations for temporal action localization in videos.
		\item  To model the interactions between proposals, we construct a graph of proposals by establishing the edges based on our valuable observations and then apply GCNs to do message aggregation among proposals.
		\item We have verified the effectiveness of our proposed method on two benchmarks. On THUMOS14 especially, our method obtains the mAP of $49.1\%$ when $tIoU=0.5$, which significantly outperforms the state-of-the-art, \ie $42.8\%$ by~\cite{chao2018rethinking}. Augmentation experiments on ActivityNet also verify the efficacy of modeling action proposal relationships.
	\end{itemize}
	
	
	\begin{figure*}[!t]
		\centering
		\includegraphics[width=\linewidth]{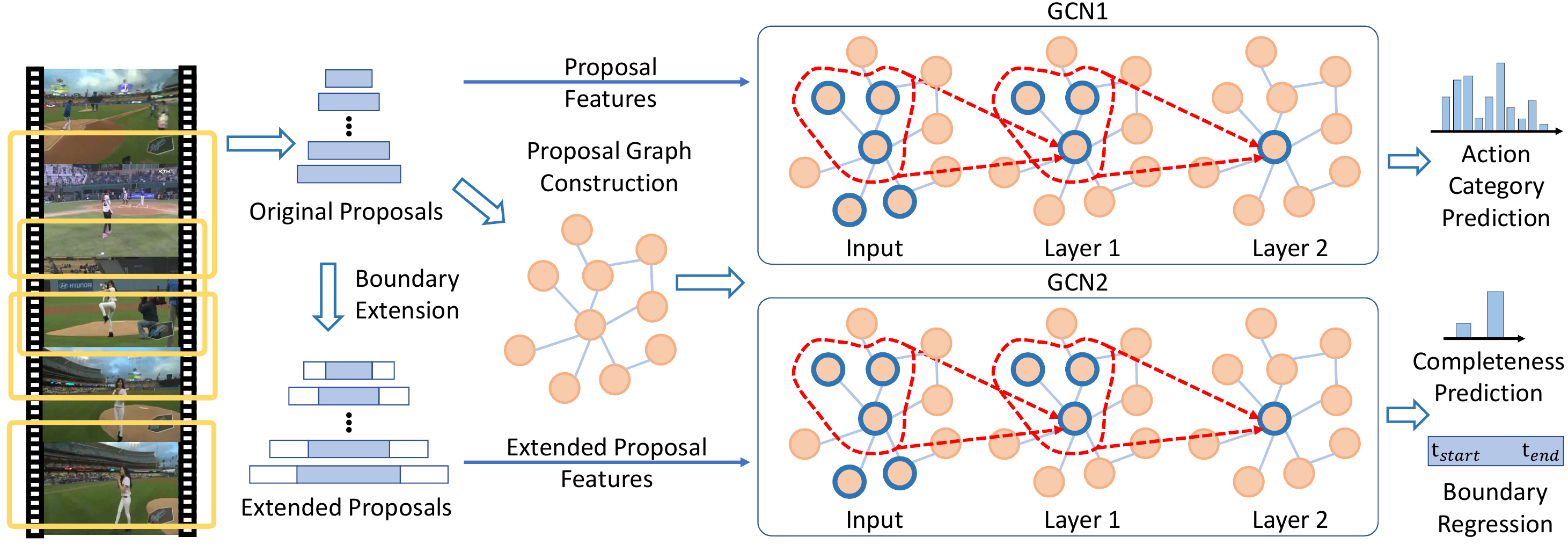}
		\caption{Schematic of our P-GCN model. Given a set of proposals from the input untrimmed video, we instantiate the nodes in the graph by each proposal. Then, edges are established among nodes to model the relations between proposals. We employ two separate GCNs on the same constructed graph with different input features (\ie, the original feature and the extended feature). Finally, P-GCN model outputs the predicted action category, completeness and boundary regression results for all proposals simultaneously.}
		\label{Fig:framework}
	\end{figure*}
	
	\section{Related work}\label{Sec:related}
	\noindent \textbf{Temporal action localization.} Recently, great progress has been achieved in deep learning~\cite{carreira2017quo,deng2018visual,guo2019auto,zhuang2018discrimination}, which facilitates the development of temporal action localization. Approaches on this task can be grouped into three categories: (1) methods performing frame or segment-level classification where the smoothing and merging steps are required to obtain the temporal boundaries~\cite{shou2017cdc,montes2016temporal,zeng2019breaking}; (2) approaches employing a two-stage framework involving proposal generation, classification and boundary refinement~\cite{shou2016temporal,xu2017r,zhao2017temporal}; (3) methods developing end-to-end architectures integrating the proposal generation and classification~\cite{yeung2016end,buch2017end,lin2017single}. 
	
	Our work is built upon the second category where the action proposals are first generated and then used to perform classification and boundary regression. Following this paradigm, Shou~\etal~\cite{shou2016temporal} propose to generate proposals from sliding windows and classify them.
	Xu~\etal~\cite{xu2017r} exploit the 3D ConvNet and propose a framework inspired by Faster R-CNN~\cite{ren2015faster}. The above methods neglect the context information of proposals, and hence some attempts have been developed to incorporate the context to enhance the proposal feature~\cite{dai2017temporal, gao2017turn, gao2017cascaded, zhao2017temporal, chao2018rethinking}. They show encouraging improvements by extracting features on the extended receptive field (\ie, boundary) of the proposal.
	Despite their success, they all process each proposal individually. In contrast, our method has considered the proposal-proposal interactions and leveraged the relations between proposals.
	
	\noindent \textbf{Graph Convolutional Networks.}
	Kipf~\etal~\cite{kipf2017semi} propose the Graph Convolutional Networks (GCNs) to define convolutions on the non-grid structures~\cite{tan2015learning}. Thanks to its effectiveness, GCNs have been successfully applied to several research areas in computer vision, such as skeleton-based action recognition~\cite{yan2018spatial}, person re-identification~\cite{shen2018person}, and video classification~\cite{wang2018video}. 
	For real-world applications, the graph can be large and directly using GCNs is inefficient. Therefore, several attempts are posed  for efficient training by virtue of the sampling strategy, such as the node-wise method SAGE~\cite{hamilton2017inductive}, layer-wise model FastGCN~\cite{chen2018fastgcn} and its layer-dependent variant AS-GCN~\cite{huang2018adaptive}. In this paper, considering the flexibility and implementability, we adopt SAGE method as the sampling strategy in our framework.

	\section{Our Approach}\label{Sec:graph}
	\subsection{Notation and Preliminaries}
	We denote an untrimmed video as $\Mat{V}=\{\Mat{I}_t\in\mathbb{R}^{H\times W\times 3}\}_{t=1}^T$, where $\Mat{I}_t$ denotes the frame at the time slot $t$ with height $H$ and width $W$.
	Within each video $\Mat{V}$,  let $\Mat{P}=\{\Mat{p}_i\mid\Mat{p}_i=(\Mat{x}_i, (t_{i,s}, t_{i,e}))\}_{i=1}^N$ be the action proposals of interest, with $t_{i,s}$ and $t_{i,e}$ being the start and end time of a proposal, respectively. In addition, given proposal $\Mat{p}_i$, let $\Mat{x}_i\in\mathcal{R}^d$ be the feature vector extracted by certain feature extractor (\eg, the I3D network~\cite{carreira2017quo}) from frames between $\Mat{I}_{t_{i,s}}$ and $\Mat{I}_{t_{i,e}}$.
	
	Let $\mathcal{G}(\mathcal{V}, \mathcal{E})$ be a graph of $N$ nodes with nodes $v_i\in\mathcal{V}$ and edge $e_{ij}=(v_i, v_j)\in\mathcal{E}$. Furthermore, let $\Mat{A}\in\mathbb{R}^{N\times N}$ be the adjacency matrix  associated with $\mathcal{G}$.
	In this paper, we seek to exploit graphs $\mathcal{G}(\mathcal{P},\mathcal{E})$ on action proposals in $\mathcal{P}$ to better model the proposal-proposal interactions in videos. Here, each action proposal is treated as a node and the edges in $\mathcal{E}$ are used to represent the relations between proposals.
	
	\subsection{General Scheme of Our Approach}
	
	In this paper, we use a proposal graph $\mathcal{G}(\mathcal{P},\mathcal{E})$ to present the relations between proposals and then apply GCN on the graph to exploit the relations and learn powerful representations for proposals. The intuition behind applying GCN is that when performing graph convolution, each node aggregates information from its neighborhoods. In this way, the feature of each proposal is enhanced by other proposals, which helps boost the detection performance eventually. 
	
	Without loss of generality, we assume the action proposals have been obtained beforehand by some methods (\eg, the TAG method in~\cite{zhao2017temporal}). In this paper, given an input video $\Mat{V}$, we seek to predict the action category $\hat{y}_i$ and temporal position $(\hat{t}_{i,s}, \hat{t}_{i,e})$ for each proposal $\Mat{p}_i$ by exploiting proposal relations. Formally, we compute
	\begin{eqnarray}
		\label{Eq:gcn-p}
		\{(\hat{y}_i, (\hat{t}_{i,s}, \hat{t}_{i,e}))\}_{i=1}^{N} = F(\mathrm{GCN}(\{\Mat{x}_i\}_{i=1}^N, \mathcal{G}(\mathcal{P},\mathcal{E})), 
	\end{eqnarray}
	where $F$ denotes any mapping functions to be learned. To exploit GCN for action localization, our paradigm takes both the proposal graph and the proposal features as input and perform graph convolution on the graph to leverage proposal relations. The enhanced proposal features (\ie, the outputs of GCN) are then used to jointly predict the category label and temporal bounding box.
	The schematic of our approach is shown in Figure~\ref{Fig:framework}. For simplicity, we denote our model as \textbf{P-GCN} henceforth.
	
	
	In the following sections, we aim to answer two questions: (1) how to construct a graph to represent the relations between proposals; (2) how to use GCN to learn representations of proposals based on the graph and facilitate the action localization.

	\subsection{Proposal Graph Construction}
	\label{Sec:construct}
	
	For the graph $\mathcal{G}(\mathcal{P},\mathcal{E})$ of each video, the nodes are instantiated as the action proposals, while the edges $\mathcal{E}$ between proposals are demanded to be characterized specifically to better model the proposal relations.
	
	One way to construct edges is linking all proposals with each other, which yet will bring in overwhelming computations for going through all proposal pairs. It also incurs redundant or noisy information for action localization, as some unrelated proposals should not be connected. In this paper, we devise a smarter approach by exploiting the temporal relevance/distance between proposals instead. Specifically, we introduce two types of edges, the contextual edges and surrounding edges, respectively. 
	
	
	
	
	
	\noindent \textbf{Contextual Edges.}
	We establish an edge between proposal $\Vec{p}_{i}$ and $\Vec{p}_{j}$ if $r(\Vec{p}_{i}, \Vec{p}_{j})> \theta_{ctx}$, where $\theta_{ctx}$ is a certain threshold.
	Here, $r(\Vec{p}_{i}, \Vec{p}_{j})$ represents the relevance between proposals and is defined by the tIoU metric, \emph{i.e.}, 
	\begin{eqnarray}
		\label{Eq:relevance}
		r(\Vec{p}_{i}, \Vec{p}_{j}) = tIoU(\Vec{p}_{i}, \Vec{p}_{j})= \frac{I(\Vec{p}_{i}, \Vec{p}_{j})}{U(\Vec{p}_{i}, \Vec{p}_{j})},
	\end{eqnarray}
	where $I(\Vec{p}_{i}, \Vec{p}_{j})$ and $U(\Vec{p}_{i}, \Vec{p}_{j})$ compute the temporal intersection and union of the two proposals, respectively. If we focus on the proposal $\Mat{p}_i$, establishing the edges by computing $r(\Vec{p}_{i}, \Vec{p}_{j})>\theta_{ctx}$ will select its neighbourhoods as those have high overlaps with it.  
	Obviously, the non-overlapping portions of the highly-overlapping neighbourhoods are able to provide rich contextual information for $\Mat{p}_i$. As already demonstrated in~\cite{dai2017temporal,chao2018rethinking}, exploring the contextual information is of great help in refining the detection boundary and increasing the detection accuracy eventually. Here, by our contextual edges, all overlapping proposals automatically share the contextual information with each other, and these information are further processed by the graph convolution.

	\noindent\textbf{Surrounding Edges.} 
	The contextual edges connect the overlapping proposals that usually correspond to the same action instance. Actually, distinct but nearby actions (including the background items) could also be correlated, and the message passing among them will facilitate the detection of each other. 
	For example in Figure~\ref{Fig:simple}, the background proposal $\Vec{p}_{4}$ will provide a guidance on identifying the action class of proposal $\Vec{p}_{1}$ (\eg, more likely to be sport action).
	To handle such kind of correlations, we first utilize $r(\Vec{p}_{i}, \Vec{p}_{j})=0$ to query the distinct proposals, and then compute the following distance 
	\begin{eqnarray}
		\label{Eq:distance}
		d(\Vec{p}_{i}, \Vec{p}_{j})=\frac{|c_{i}-c_{j}|}{U(\Vec{p}_{i}, \Vec{p}_{j})},
	\end{eqnarray}
	to add the edges between nearby proposals if $d(\Vec{p}_{i}, \Vec{p}_{j}) < \theta_{sur}$, where $\theta_{sur}$ is a certain threshold. In Eq.~\eqref{Eq:distance}, $c_{i}$ (or $c_{j}$) represents the center coordinate of $\Mat{p}_i$ (or $\Mat{p}_j$). 
	As a complement of the contextual edges, the surrounding edges enable the message to pass across distinct action instances and thereby provides more temporal cues for the detection. 
	
	
	\subsection{Graph Convolution for Action Localization} \label{Sec:gcn}
	
	Given the constructed graph, we apply the GCN to do action localization.
	We build $K$-layer graph convolutions in our implementation. 
	Specifically for the $k$-th layer ($1\leq k\leq K$), the graph convolution is implemented by
	\begin{equation} 
		\label{Eq:gcn}
		\Mat{X}^{(k)} = \Mat{A}\Mat{X}^{(k-1)}\Mat{W}^{(k)}.
	\end{equation}
	Here, $\Mat{A}$ is the adjacency matrix; $\textbf{W}^{(k)}\in\mathbb{R}^{d_k\times d_k}$ is the parameter matrix to be learned; $\Mat{X}^{(k)}\in\mathbb{R}^{N \times d_k}$ are the hidden features for all proposals at layer $k$; $\Mat{X}^{(0)}\in\mathbb{R}^{N \times d}$ are the input features.

	We apply an activation function (\ie, ReLU) after each convolution layer before the features are forwarded to the next layer. In addition, our experiments find it more effective by further concatenating the hidden features with the input features in the last layer, namely, 
	\begin{eqnarray}
		\label{Eq:layer-wise}
		\Mat{X}^{(K)} := \Mat{X}^{(K)} \| \Mat{X}^{(0)},
	\end{eqnarray}
	where $\|$ denotes the concatenation operation.   
	
	Joining the previous work~\cite{zhao2017temporal}, we find that it is beneficial to predict the action label and temporal boundary separately by virtue of two GCNs---one conducted on the original proposal features $\Mat{x}_i$ and the other one on the extended proposal features $\Mat{x'}_i$. The first GCN is formulated as 
	\begin{equation}
		\label{Eq:1gcn}
		\{\hat{y}_i\}_{i=1}^{N} =  \mathrm{softmax}(FC_1(GCN_1(\{\Mat{x}_i\}_{i=1}^N, \mathcal{G}(\mathcal{P},\mathcal{E})))), \\
	\end{equation}
	where we apply a Fully-Connected (FC) layer with soft-max operation on top of $GCN_1$ to predict the action label $\hat{y}_i$. The second GCN can be formulated as 
	\begin{eqnarray}
		\label{Eq:2gcn_1}
		\{(\hat{t}_{i,s}, \hat{t}_{i,e})\}_{i=1}^{N} = FC_2(GCN_2(\{\Mat{x'}_i\}_{i=1}^N, \mathcal{G}(\mathcal{P},\mathcal{E}))), \\
		\label{Eq:2gcn_2}
		\{\hat{c}_i\}_{i=1}^{N} = FC_3(GCN_2(\{\Mat{x'}_i\}_{i=1}^N, \mathcal{G}(\mathcal{P},\mathcal{E}))),
	\end{eqnarray}
	where the graph structure $\mathcal{G}(\mathcal{P},\mathcal{E})$ is the same as that in Eq.~\eqref{Eq:1gcn} but the input proposal feature is different. The extended feature $\Mat{x'}_i$ is attained by first extending the temporal boundary of $\Mat{p}_i$ with $\frac{1}{2}$ of its length on both the left and right sides and then extracting the feature within the extended boundary. Here, we adopt two FC layers on top of $GCN_2$, one for predicting the boundary $(\hat{t}_{i,s}, \hat{t}_{i,e})$ and the other one for predicting the completeness label $\hat{c}_i$, which indicates whether the proposal is complete or not. It has been demonstrated by~\cite{zhao2017temporal} that, incomplete proposals that have low tIoU with the ground-truths could have high classification score, and thus it will make mistakes when using the classification score alone to rank the proposal for the mAP test; further applying the completeness score enables us to avoid this issue.
	
	\noindent\textbf{Adjacency Matrix.}  
	In Eq.~\eqref{Eq:gcn}, we need to compute the adjacency matrix $\Mat{A}$.
	Here, we design the adjacency matrix by
	assigning specific weights to edges. For example, we can apply the cosine similarity to estimate the weights of edge $e_{ij}$ by
	
	\begin{equation}
		\label{Eq:adjacent_mat}
		A_{ij}=\frac{\Vec{x}_{i}^{T}\Vec{x}_{j}}{\|\Vec{x}_{i}\|_{2}\cdot\|\Vec{x}_{j}\|_{2}}.
	\end{equation}
	In the above computation, we compute $A_{ij}$ relying on the feature vector $\Vec{x}_{i}$. We can also map the feature vectors 
	into an embedding space using a learnable linear mapping function as in \cite{wang2018non}
	before the cosine computation. We leave the discussion in our experiments.
	
	
	\subsection{Efficient Training by Sampling}
	\label{Sec:training}
	
	Typical proposal generation methods usually produce thousands of proposals for each video. 
	Applying the aforementioned graph convolution (Eq.~\eqref{Eq:gcn}) on all proposals demands hefty computation and memory footprints. To accelerate the training of GCNs, several approaches~\cite{chen2018fastgcn,huang2018adaptive,hamilton2017inductive} have been proposed based on neighbourhood sampling. Here, we adopt the SAGE method~\cite{hamilton2017inductive} in our method for its flexibility. 
	
	The SAGE method uniformly samples the fixed-size neighbourhoods of each node layer-by-layer in a top-down passway. In other words, the nodes of the $(k-1)$-th layer are formulated as the sampled neighbourhoods of the nodes in the $k$-th layer. After all nodes of all layers are sampled, SAGE performs the information aggregation in a bottom-up fashion. Here we specify the aggregation function to be a sampling form of Eq.~\eqref{Eq:gcn}, namely,
	\begin{equation} 
		\label{Eq:gcn-sampling}
		\Mat{x}_i^{(k)} = \left(\frac{1}{N_s} \sum_{j=1}^{N_s} A_{ij}\Mat{x}_j^{(k-1)} + \Mat{x}_i^{(k-1)}\right)\Mat{W}^{(k)},
	\end{equation}
	where node $j$ is sampled from the neighbourhoods of node $i$, \ie, $j\in\mathcal{N}(i)$; $N_s$ is the sampling size and is much less than the total number $N$. The summation in Eq.~\eqref{Eq:gcn-sampling} is further normalized by $N_s$, which empirically makes the training more stable. Besides, we also enforce the self addition of its feature for node $i$ in Eq.~\eqref{Eq:gcn-sampling}.
	We do not perform any sampling when testing. For better readability, Algorithm~\ref{Alg:forward} depicts the algorithmic Flow of our method.

	\begin{algorithm}[!bt]
		\caption{The training process of P-GCN model.}
		\begin{flushleft}
			\textbf{Input:} Proposal set $\Mat{P}=\{\Mat{p}_i\mid\Mat{p}_i=(\Mat{x}_i, (t_{i,s}, t_{i,e}))\}_{i=1}^N$;
			original proposal features $\{\Mat{x}_{i}^{(0)}\}_{i=1}^N$;
			extended proposal features $\{\Mat{x'}_{i}^{(0)}\}_{i=1}^N$;
			graph depth $K$; sampling size $N_s$
			\vspace{-0.3cm}
		\end{flushleft}
		\begin{flushleft}
			\textbf{Parameter:} Weight matrices $\textbf{W}^{(k)}$, $\forall k \in \{1,\dots,K\}$
			\vspace{-0.3cm}
		\end{flushleft}
		\begin{algorithmic}[1]
			\State instantiate the nodes by the proposals $\Mat{p}_i$, $\forall \Mat{p}_i \in \Mat{P}$
			\State establish edges between nodes
			\State obtain a proposal graph $\mathcal{G}(\mathcal{P}, \mathcal{E})$
			\State calculate adjacent matrix using Eq.~\eqref{Eq:adjacent_mat}
			\While {not converges}
			\For {$k=1\dots K$}
			\For {$p \in \mathcal{P}$}
			\State sample $N_s$ neighborhoods of $p$
			\State aggregate information using Eq.~\eqref{Eq:gcn-sampling}
			\EndFor
			\EndFor
			\State predict action categories $\{\hat{y}_i\}_{i=1}^{N}$ using Eq.~\eqref{Eq:1gcn}
			\State perform boundary regression using Eq.~\eqref{Eq:2gcn_1}
			\State predict completeness $\{\hat{c}_i\}_{i=1}^{N}$ using Eq.~\eqref{Eq:2gcn_2}
			\EndWhile
			\vspace{-0.3cm}
		\end{algorithmic}
		\begin{flushleft}
			\textbf{Output:} Trained P-GCN model
		\end{flushleft}
		\label{Alg:forward}
		\vspace{-0.4cm}
	\end{algorithm}


	\section{Experiments}
	
	\subsection{Datasets}
	\textbf{THUMOS14 \cite{jiang2014thumos}} is a standard benchmark for action localization. 
	Its training set known as the UCF-101 dataset consists of 13320 videos. The validation, testing and background set contain 1010, 1574 and 2500 untrimmed videos, respectively. Performing action localization on this dataset is challenging since each video has more than 15 action instances and its 71\% frames are occupied by background items.
	Following the common setting in \cite{jiang2014thumos}, we apply 200 videos in the validation set for training and conduct evaluation on the 213 annotated videos from the testing set. 
	
	\textbf{ActivityNet \cite{caba2015activitynet}} is another popular benchmark for action localization on untrimmed videos. We evaluate our method on ActivityNet v1.3, which contains around 10K training videos and 5K validation videos corresponded to 200 different activities. Each video has an average of 1.65 action instances. Following the standard practice, we train our method on the training videos and test it on the validation videos. In our experiments, we contrast our method with the state-of-the-art methods on both THUMOS14 and ActivityNet v1.3, and perform ablation studies on THUMOS14.
	
	
	\subsection{Implementation details}
	
	\noindent \textbf{Evaluation Metrics.} We use mean Average Precision (mAP) as the evaluation metric. 
	A proposal is considered to be correct if its temporal IoU with the ground-truth instance is larger than a certain threshold and the predicted category is the same as this ground-truth instance.
	On THUMOS14, the tIOU thresholds are chosen from $\{0.1, 0.2, 0.3, 0.4, 0.5\}$; on ActivityNet v1.3, the IoU thresholds are from $\{0.5, 0.75, 0.95\}$, and we also report the average mAP of the IoU thresholds between 0.5 and 0.95 with the step of $0.05$.
	
	\noindent \textbf{Features and Proposals.}
	Our model is implemented under the two-stream strategy~\cite{simonyan2014two}: RGB frames and optical flow fields.
	We first uniformly divide each input video into 64-frame segments. We then use a two-stream Inflated 3D ConvNet (I3D) model pre-trained on Kinetics~\cite{carreira2017quo} to extract the segment features. In detail, the I3D model takes as input the RGB/optical-flow segment and outputs a 1024-dimensional feature vector for each segment. Upon the I3D features, we further apply max pooling across segments to obtain one 1024-dimensional feature vector for each proposal that is obtained by the BSN method~\cite{lin2018bsn}. Note that we do not finetune the parameters of the I3D model in our training phase. 
	Besides the I3D features and BSN proposals, our ablation studies in \textsection~\ref{Sec:ablation} also explore other types of features (\eg 2-D features~\cite{lin2018bsn}) and proposals  (\eg TAG proposals~\cite{zhao2017temporal}).

	\noindent \textbf{Proposal Graph Construction.}
	We construct the proposal graph by fixing the values of $\theta_{ctx}$ as 0.7 and  $\theta_{sur}$ as 1 for both streams. More discussions on choosing the values of $\theta_{ctx}$ and $\theta_{sur}$ could be found in the supplementary material. We adopt 2-layer GCN since we observed no clear improvement with more than 2 layers but the model complexity is increased. For more efficiency, we choose $N_{s}=4$ in Eq.~\eqref{Eq:gcn-sampling} for neighbourhood sampling unless otherwise specified.

	\noindent \textbf{Training.} 
	The initial learning rate is 0.001 for the RGB stream and 0.01 for the Flow stream. 
	During training, the learning rates will be divided by 10 every 15 epochs. The dropout ratio is 0.8.
	The classification $\hat{y}_i$ and completeness $\hat{c}_i$ are trained with the cross-entropy loss and the hinge loss, respectively. The regression term $(\hat{t}_{i,s}, \hat{t}_{i,e})$ is trained with the smooth $L_{1}$ loss. More training details can be found in the supplementary material.


	
	
	\noindent \textbf{Testing.} 
	We do not perform neighbourhood sampling (\emph{i.e.} Eq.~\eqref{Eq:gcn-sampling}) for testing. The predictions of the RGB and Flow steams are fused using a ratio of 2:3.
	We multiply the classification score with the completeness score as the final score for calculating mAP.
	We then use Non-Maximum Suppression (NMS) to obtain the final predicted temporal proposals for each action class separately.
	We use 600 and 100 proposals per video for computing mAPs on THUMOS14 and ActivityNet v1.3, respectively.

	\subsection{Comparison with state-of-the-art results}
	\noindent \textbf{THUMOS14.}
	Our P-GCN model is compared with the state-of-the-art methods in Table \ref{Tab:thumos}. The P-GCN model reaches the highest mAP over all thresholds, implying that our method can recognize and localize actions much more accurately than any other method. Particularly, our P-GCN model outperforms the previously best method (\emph{i.e.} TAL-Net~\cite{chao2018rethinking}) by 6.3\% absolute improvement and the second-best result~\cite{lin2018bsn} by more than 12.2\%, when $tIoU=0.5$.
	
	\begin{table}[!tb]
		\centering
		\caption{Action localization results on THUMOS14, measured by mAP (\%) at different tIoU thresholds $\alpha$.}
		\vspace{0.1cm}
		\begin{tabular}{lccccccc}
			\hline
			tIoU                         & 0.1           & 0.2           & 0.3           & 0.4           & 0.5                     \\ \hline
			Oneata \etal \cite{oneata2014lear}       & 36.6          & 33.6          & 27.0          & 20.8          & 14.4                  \\
			Wang \etal \cite{wang2014action}         & 18.2          & 17.0          & 14.0          & 11.7          & 8.3                     \\
			Caba \etal \cite{caba2016fast}           & -             & -            & -             & -             & 13.5                 \\
			Richard \etal \cite{richard2016temporal} & 39.7          & 35.7          & 30.0          & 23.2          & 15.2                 \\
			Shou \etal \cite{shou2016temporal}       & 47.7          & 43.5          & 36.3          & 28.7          & 19.0                   \\
			Yeung \etal \cite{yeung2016end}          & 48.9          & 44.0          & 36.0          & 26.4          & 17.1              \\
			Yuan \etal \cite{yuan2016temporal}       & 51.4          & 42.6          & 33.6          & 26.1          & 18.8                   \\
			Escorcia \etal  \cite{escorcia2016daps}  & -             & -             & -             & -             & 13.9                   \\
			Buch \etal  \cite{buch2017sst}           & -             & -             & 37.8          & -             & 23.0                    \\
			Shou \etal \cite{shou2017cdc}            & -             & -            & 40.1          & 29.4          & 23.3                    \\
			Yuan \etal \cite{Yuan2017}       & 51.0          & 45.2          & 36.5          & 27.8          & 17.8                    \\
			Buch \etal \cite{buch2017end}            & -             & -             & 45.7          & -             & 29.2                    \\
			Gao \etal \cite{gao2017cascaded}         & 60.1          & 56.7          & 50.1          & 41.3          & 31.0                    \\
			Hou \etal \cite{hou2017real}             & 51.3          & -             & 43.7          & -            & 22.0                      \\
			Dai \etal \cite{dai2017temporal}         & -             & -            & -             & 33.3          & 25.6                  \\
			Gao \etal \cite{gao2017turn}             & 54.0          & 50.9          & 44.1          & 34.9          & 25.6                   \\
			Xu \etal \cite{xu2017r}                  & 54.5          & 51.5          & 44.8          & 35.6          & 28.9                     \\
			Zhao \etal \cite{zhao2017temporal}       & 66.0          & 59.4          & 51.9          & 41.0          & 29.8                   \\ 
			Lin \etal \cite{lin2018bsn}              & -             & -             & 53.5          & 45.0          & 36.9                  \\
			Chao \etal \cite{chao2018rethinking}     & 59.8          & 57.1          & 53.2          & 48.5          & 42.8          \\ \hline
			P-GCN                          & \textbf{69.5} & \textbf{67.8} & \textbf{63.6} & \textbf{57.8} & \textbf{49.1}  \\ \hline
		\end{tabular}
		\label{Tab:thumos}
	\end{table}
	
	\begin{table}[!t]
		\centering
		\caption{Action localization results on ActivityNet v1.3 (val), measured by mAP (\%) at different tIoU thresholds and the average mAP of IoU thresholds from 0.5 to 0.95. (*) indicates the method that uses the external video labels from UntrimmedNet \cite{wang2017untrimmed}.}
		\vspace{0.1cm}
		\begin{tabular}{lccc|c}
			\hline
			tIoU                         & 0.5           & 0.75          & 0.95           & Average              \\ \hline
			Singh \etal  \cite{singh2016untrimmed}        & 34.47        & -             & -               & -                 \\
			Wang \etal   \cite{wang2016uts}        & 43.65        & -             & -               & -                 \\
			Shou \etal \cite{shou2017cdc}         & 45.30             & 26.00             & 0.20          & 23.80                      \\
			Dai \etal \cite{dai2017temporal}        & 36.44             & 21.15             & 3.90             & -                        \\
			Xu \etal \cite{xu2017r}         & 26.80          & -          & -          & -                         \\
			Zhao \etal \cite{zhao2017temporal}        & 39.12 & 23.48          & 5.49           & 23.98                       \\ 
			Chao \etal  \cite{chao2018rethinking}          & 38.23          & 18.30          & 1.30          & 20.22         \\  
			P-GCN                & \textbf{42.90}         & \textbf{28.14} & 2.47 & \textbf{26.99}  \\ \hline 
			Lin \etal  \cite{lin2018bsn} *    & 46.45        & 29.96          & 8.02        & 30.03 \\ 
			P-GCN* & \textbf{48.26}  & \textbf{33.16} & 3.27 & \textbf{31.11} \\
			\hline
		\end{tabular}
		\label{Tab:anet}
	\end{table}

	\noindent \textbf{ActivityNet v1.3.} Table \ref{Tab:anet} reports the action localization results of various methods. Regarding the average mAP, P-GCN outperforms SSN~\cite{zhao2017temporal}, CDC~\cite{shou2017cdc}, and TAL-Net~\cite{chao2018rethinking} by 3.01\%, 3.19\%, and 6.77\%, respectively. 
	We observe that the method by Lin~\etal \cite{lin2018bsn} (called LIN below) performs promisingly on this dataset. Note that LIN is originally designed for generating class-agnostic proposals, and thus relies on external video-level action labels (from UntrimmedNet \cite{wang2017untrimmed}) for action localization. 
	In contrast, our method is self-contained and is able to perform action localization without any external label. 
	Actually, P-GCN can still be modified to take external labels into account. 
	To achieve this, we assign the top-2 video-level classes predicted by UntrimmedNet to all the proposals in that video.
	We provide more details about how to involve external labels in P-GCN in the supplementary material. As summarized in Table~\ref{Tab:anet},  our enhanced version P-GCN* consistently outperforms LIN, hence demonstrating the effectiveness of our method under the same setting.
	
	\begin{figure}[!tb]
		\centering
		\includegraphics[width=\linewidth]{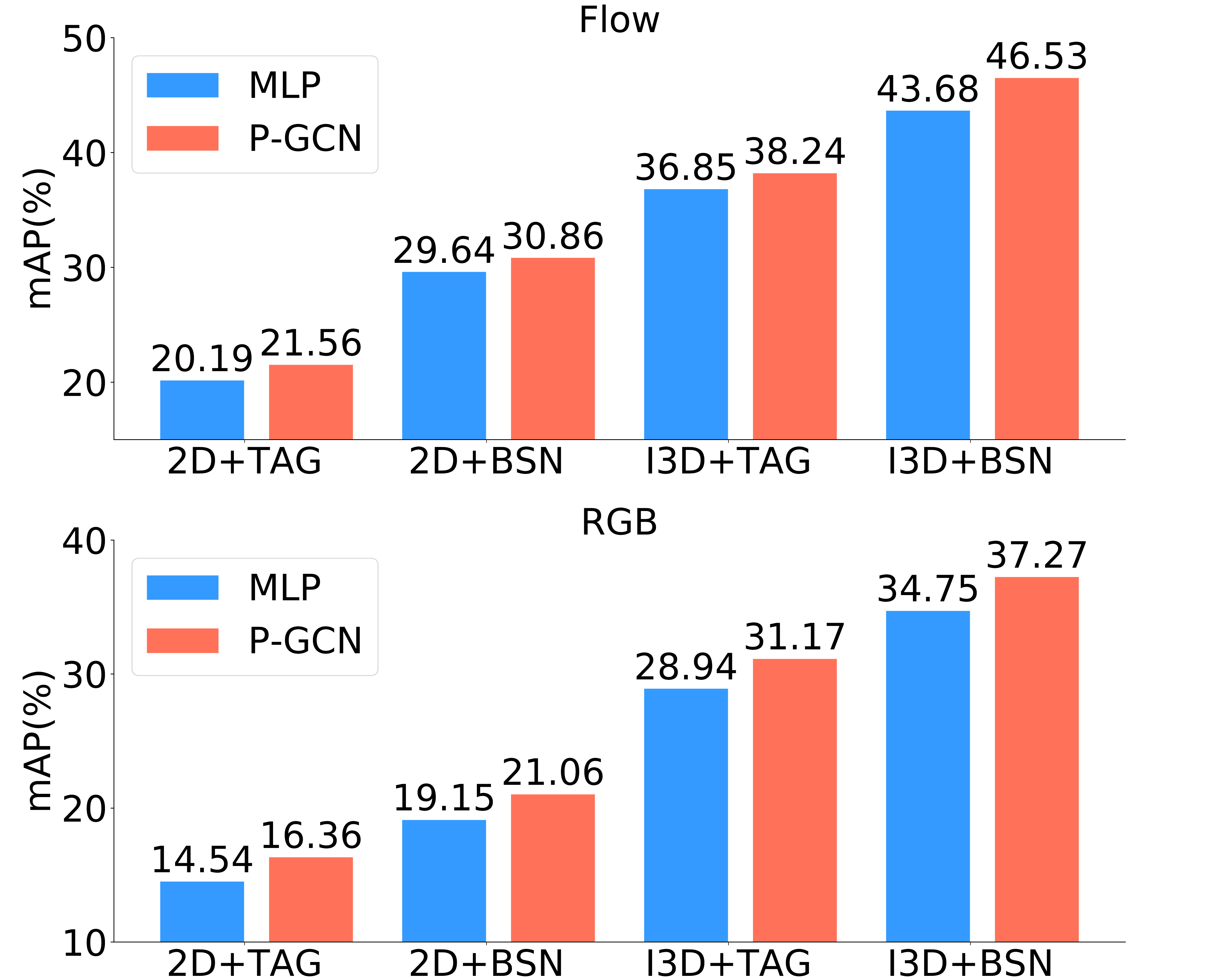}
		\caption{Action localization results on THUMOS14 with different backbones, measured by mAP@tIoU=0.5.}
		\vspace{-0.3cm}
		\label{Fig:instantiations}
	\end{figure}

	\section{Ablation Studies}
	In this section, we will perform complete and in-depth ablation studies to evaluate the impact of each component of our model. More details about the structures of baseline methods (such as MLP and MP) can be found in the supplementary material.
	\label{Sec:ablation}
	\subsection{How do the proposal-proposal relations help?}
	\label{Sec:5.1}
	As illustrated in \textsection~\ref{Sec:gcn}, we apply two GCNs for action classification and boundary regression separately. 
	Here, we implement the baseline with a 2-layer MultiLayer-Perceptron (MLP). The MLP baseline shares the same structure as GCN except that we remove the adjacent matrix $\Mat{A}$ in Eq.~\eqref{Eq:gcn}. To be specific, for the $k$-th layer, the propagation in Eq.~\eqref{Eq:gcn} becomes $\Mat{X}^{k}=\Mat{X}^{k-1}\Mat{W}^k$, where $\Mat{W}^k$ are the trainable parameters. 
	Without using $\Mat{A}$, MLP processes each proposal feature independently. By comparing the performance of MLP with GCN, we can justify the importance of message passing along proposals. 
	To do so, we replace each GCN with an MLP and have the following variants of our model including: (1) \textbf{MLP$_1$ + GCN$_2$} where GCN$_1$ is replaced; (2) \textbf{GCN$_1$ + MLP$_2$} where GCN$_2$ is replaced; and (3) \textbf{MLP$_1$ + MLP$_2$} where both GCNs are replaced.
	Table \ref{Tab:twographs} reads that all these variants decrease the performance of our model, thus verifying the effectiveness of GCNs for both action classification and boundary regression. 
	Overall, our model P-GCN significantly outperforms the MLP protocol (\ie \textbf{MLP$_1$ + MLP$_2$}), validating the importance of considering proposal-proposal relations in temporal action localization.

	\subsection{How does the graph convolution help?}
	Besides graph convolutions, performing mean pooling among proposal features is another way to enable information dissemination between proposals. We thus conduct another baseline by first adopting MLP on the proposal features and then conducting mean pooling on the output of MLP over adjacent proposals. The adjacent connections are formulated by using the same graph as GCN. We term this baseline as MP below.
	Similar to the setting in \textsection~\ref{Sec:5.1}, we have three variants of our model including: (1) \textbf{MP$_1$ + MP$_2$}; (2) \textbf{MP$_1$ + GCN$_2$}; and (3) \textbf{GCN$_1$ + MP$_2$}. We report the results in Table \ref{Tab:mean-pooling}.  Our P-GCN outperforms all MP variants, demonstrating the superiority of graph convolution over mean pooling on capturing between-proposal connections.
	The protocol \textbf{MP$_1$ + MP$_2$} in Table~\ref{Tab:mean-pooling} performs better than \textbf{MLP$_1$ + MLP$_2$} in Table~\ref{Tab:twographs}, which again reveals the benefit of modeling the proposal-proposal relations, even we pursue it using the naive mean pooling.
	
	\begin{table}[!tb]
		\centering
		\caption{Comparison between our P-GCN model and MLP on THUMOS14, measured by mAP (\%).}
		\vspace{0.1cm}
		\begin{tabular}{l|cc|cc}
			\hline
			mAP@tIoU=0.5                & RGB     & Gain      & Flow     & Gain        \\ \hline
			MLP$_1$ + MLP$_2$        & 34.75      & -       & 43.68  & -\\
			MLP$_1$ + GCN$_2$  & 35.94   & 1.19       & 44.59 &0.91 \\
			GCN$_1$ + MLP$_2$  & 35.82   & 1.07        & 45.26  & 1.58 \\
			P-GCN (GCN$_1$ + GCN$_2$)   & \textbf{37.27}   & \textbf{2.52}       & \textbf{46.53}  &  \textbf{2.85}\\ \hline
		\end{tabular}
		\label{Tab:twographs}
	\end{table} 
	
	\begin{table}[!tb]
		\centering
		\caption{Comparison between our P-GCN model and mean-pooling (MP) on THUMOS14, measured by mAP (\%).}
		\vspace{0.1cm}
		\begin{tabular}{l|cc|cc}
			\hline
			mAP@tIoU=0.5                & RGB     & Gain      & Flow     & Gain        \\ \hline
			MP$_1$ + MP$_2$        & 35.32      & -       & 43.97  & -\\
			MP$_1$ + GCN$_2$  & 36.50   & 1.18       & 45.78 & 1.81 \\
			GCN$_1$ + MP$_2$  & 36.22   & 0.90        & 44.42  & 0.45 \\
			P-GCN (GCN$_1$ + GCN$_2$)   & \textbf{37.27}   & \textbf{1.95}       & \textbf{46.53}  &  \textbf{2.56}\\ \hline
		\end{tabular}
		\label{Tab:mean-pooling}
	\end{table}

	\begin{table}[!tb]
		\caption{Comparison of different types of edge functions on THUMOS14, measured by mAP (\%).}
		\vspace{0.1cm}
		\centering
		\begin{tabular}{l|cc}
			\hline
			mAP@tIoU=0.5                & RGB      & Flow      \\ \hline
			MLP         & 34.75     & 43.68 \\
			P-GCN(cos-sim)   & 35.55    & 44.83   \\
			P-GCN(cos-sim, self-add)    & 37.27  & 46.53 \\ 
			P-GCN(embed-cos-sim, self-add)    & 36.81   & 46.89 \\ \hline 
		\end{tabular}
			 		\vspace{-0.1cm}
		\label{Tab:edge}
	\end{table}
	
	\subsection{Influences of different backbones}
	Our framework is general and compatible with different backbones (\ie, proposals and features). Beside the backbones applied above, we further perform experiments on TAG proposals~\cite{zhao2017temporal} and 2D features~\cite{lin2018bsn}.
	We try different combinations: (1) BSN+I3D; (2) BSN+2D; (3) TAG+I3D; (4) TAG+2D, and report the results of MLP and P-GCN in Figure \ref{Fig:instantiations}. In comparison with MLP, our P-GCN leads to significant and consistent improvements in all types of features and proposals. These results conclude that, our method is generally effective and is not limited to the specific feature or proposal type. 

	\subsection{The weights of edge and self-addition}
	
	We have defined the weights of edges in Eq.~\eqref{Eq:adjacent_mat}, where the cosine similarity (cos-sim) is applied. This similarity can be further extended by first embedding the features before the cosine computation. We call the embedded version as embed-cos-sim, and compare it with cos-sim in Table \ref{Tab:edge}. 
	No obvious improvement is attained by replacing cos-sim with embed-cos-sim (the mAP difference between them is less than $0.4\%$). Eq.~\eqref{Eq:gcn-sampling} has considered the self-addition of the node feature. We also investigate the importance of this term in Table~\ref{Tab:edge}. It suggests that the self-addition leads to at least 1.7\% absolute improvements on both RGB and Flow streams.
	
	\subsection{Is it necessary to consider two types of edges?}
	
	To evaluate the necessity of formulating two types of edges, we perform experiments on two variants of our P-GCN, each of which considers only one type of edge in the graph construction stage. As expected, the result in Table~\ref{Tab:surrounding} drops remarkably when either kind of edge is removed. 
	Another crucial point is that our P-GCN still boosts MLP when only the surrounding edges are remained. The rationale behind this could be that, actions in the same video are correlated and exploiting the surrounding relation will enable more accurate action classification. 
	
	\begin{table}[!tb]
		\centering
		\caption{Comparison of two types of edge on THUMOS14, measured by mAP (\%).}
		\vspace{0.1cm}
		\begin{tabular}{l|cc|cc}
			\hline
			mAP@tIoU=0.5                & RGB     & Gain      & Flow     & Gain        \\ \hline
			w/ both edges (P-GCN)    & 37.27   & -       & 46.53  &  - \\
			w/o surrounding edges   & 35.84   & -1.43      & 45.89  & -0.64 \\ 
			w/o contextual edges   & 36.81   &  -0.46     & 45.57   & -0.96 \\ 
			w/o both edges (MLP)             & 34.75   & -2.52     & 43.68      & -2.85 \\ \hline
		\end{tabular}
		\label{Tab:surrounding}
	\end{table}
	
	\begin{table}[!t]
		\centering
		\tabcolsep 5pt 
		\caption{Comparison of different sampling size and training time for each iteration on THUMOS14, measured by mAP@tIoU=0.5.}
		\vspace{0.1cm}
		\begin{tabular}{c|cccccc}
			\hline
			$N_s$  & 1 & 2 & 3 & 4 & 5 & 10 \\ \hline
			RGB  & 36.0  & 36.92 & 35.68 & \textbf{37.27} & 36.11 & 36.37\\
			Flow & 46.15 & 45.06 & 45.13 & \textbf{46.53} & 46.28 & 46.14\\
			Time(s) & 0.10 & 0.23 & 0.33 & 0.41 & 0.48 & 1.72 \\ \hline
		\end{tabular}
			 		\vspace{-0.2cm}
		\label{Tab:sampling}
	\end{table}
	
	\subsection{The efficiency of our sampling strategy}
	
	We train P-GCN efficiently based on the neighbourhood sampling in Eq.~\eqref{Eq:gcn-sampling}. Here, we are interested in how the sampling size $N_{s}$ affects the final performance. Table \ref{Tab:sampling} reports the testing mAPs corresponded to different $N_s$ varying from 1 to 5 (and also 10). The training time per iteration is also added in Table \ref{Tab:sampling}. 
	We observe that when $N_{s}=4$ the model achieves higher mAP than the full model (\ie, $N_{s}=10$) while reducing 76\% of training time for each iteration. This is interesting, as sampling fewer nodes even yields better results. We conjecture that, the neighbourhood sampling could bring in more stochasticity and guide our model to escape from the local minimal during training, thus delivering better results.
	
	\begin{figure}[!t]
		\centering
		\includegraphics[width=\linewidth]{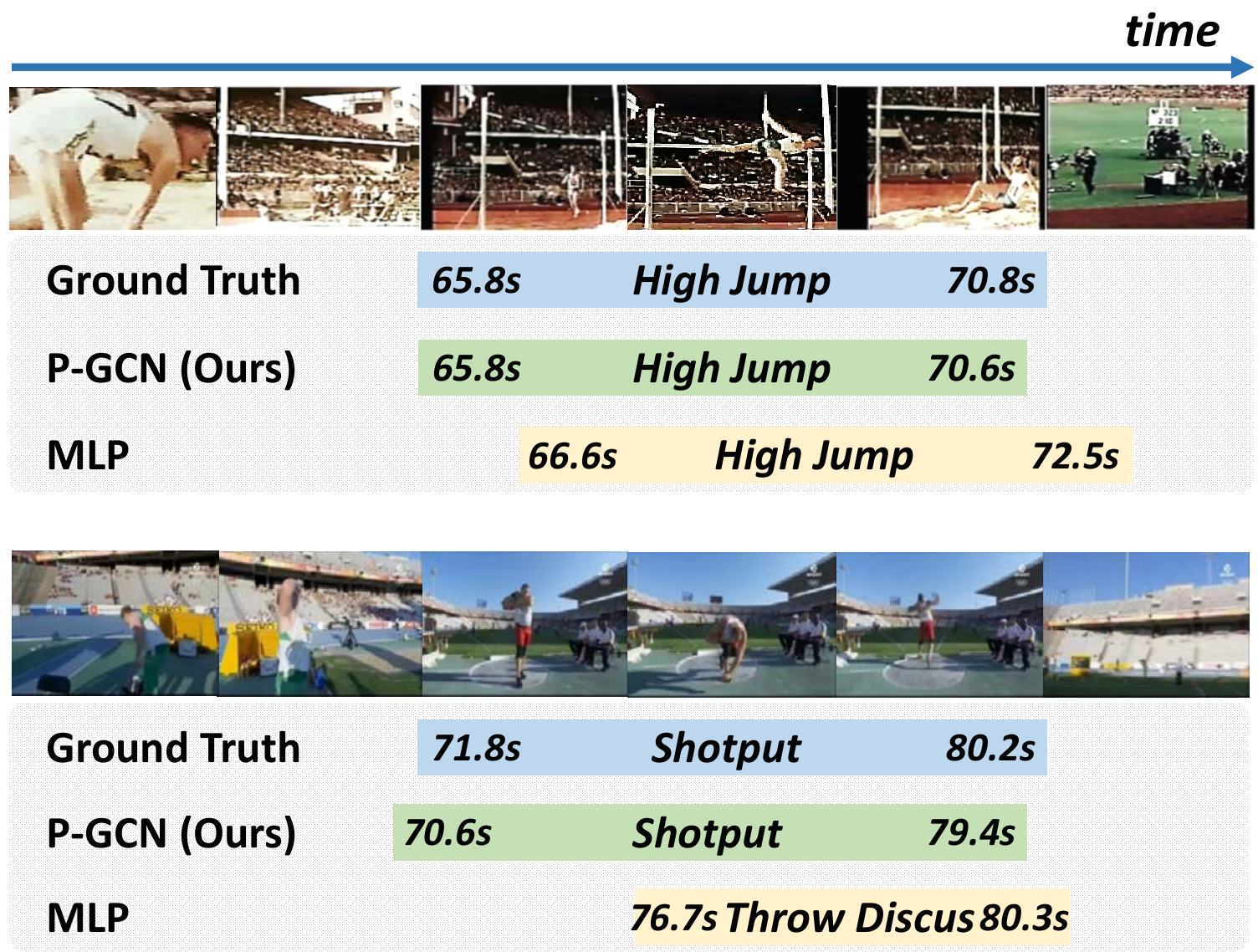}
		\caption{Qualitative results on THUMOS14 dataset.}
			 	\vspace{-0.3cm}
		\label{Fig:qualitative}
	\end{figure}
	
	\subsection{Qualitative Results}
	
	Given the significant improvements, we also attempt to find out in what cases our P-GCN model improves over MLP. We visualize the qualitative results on THUMOS14 in Figure ~\ref{Fig:qualitative}. In the top example, both MLP and our P-GCN model are able to predict the action category correctly, while P-GCN predicts a more precise temporal boundary. In the bottom example, due to similar action characteristic and context, MLP predicts the action of ``Shotput'' as ``Throw Discus''. Despite such challenge, P-GCN still correctly predicts the action category, demonstrating the effectiveness of our method. More qualitative results could be found in the supplementary material.
	
	\section{Conclusions}
	
	In this paper, we have exploited the proposal-proposal interaction to tackle the task of temporal action localization. By constructing a graph of proposals and applying GCNs to message passing, our P-GCN model outperforms the state-of-the-art methods by a large margin on two benchmarks, \ie, THUMOS14 and ActivithNet v1.3. It would be interesting to extend our P-GCN for object detection in image and we leave it for our future work.
	
	\footnotesize 
	{\flushleft \bf Acknowledgements}. This work was partially supported by National Natural Science Foundation of China (NSFC) 61602185, 61836003 (key project),
	Program for Guangdong Introducing Innovative and Enterpreneurial Teams 2017ZT07X183, Guangdong Provincial Scientific and Technological Funds under Grants 2018B010107001, and Tencent AI Lab Rhino-Bird Focused Research Program (No. JR201902).
	

\newpage
\normalsize
\setcounter{section}{0}
\renewcommand\thesection{\Alph{section}}
\setcounter{figure}{0}
\renewcommand\thefigure{\Alph{figure}}
\setcounter{table}{0}
\renewcommand\thetable{\Alph{table}}
\section{Proposal Features}

\begin{figure}[!h]
	\centering
	\includegraphics[width=\linewidth]{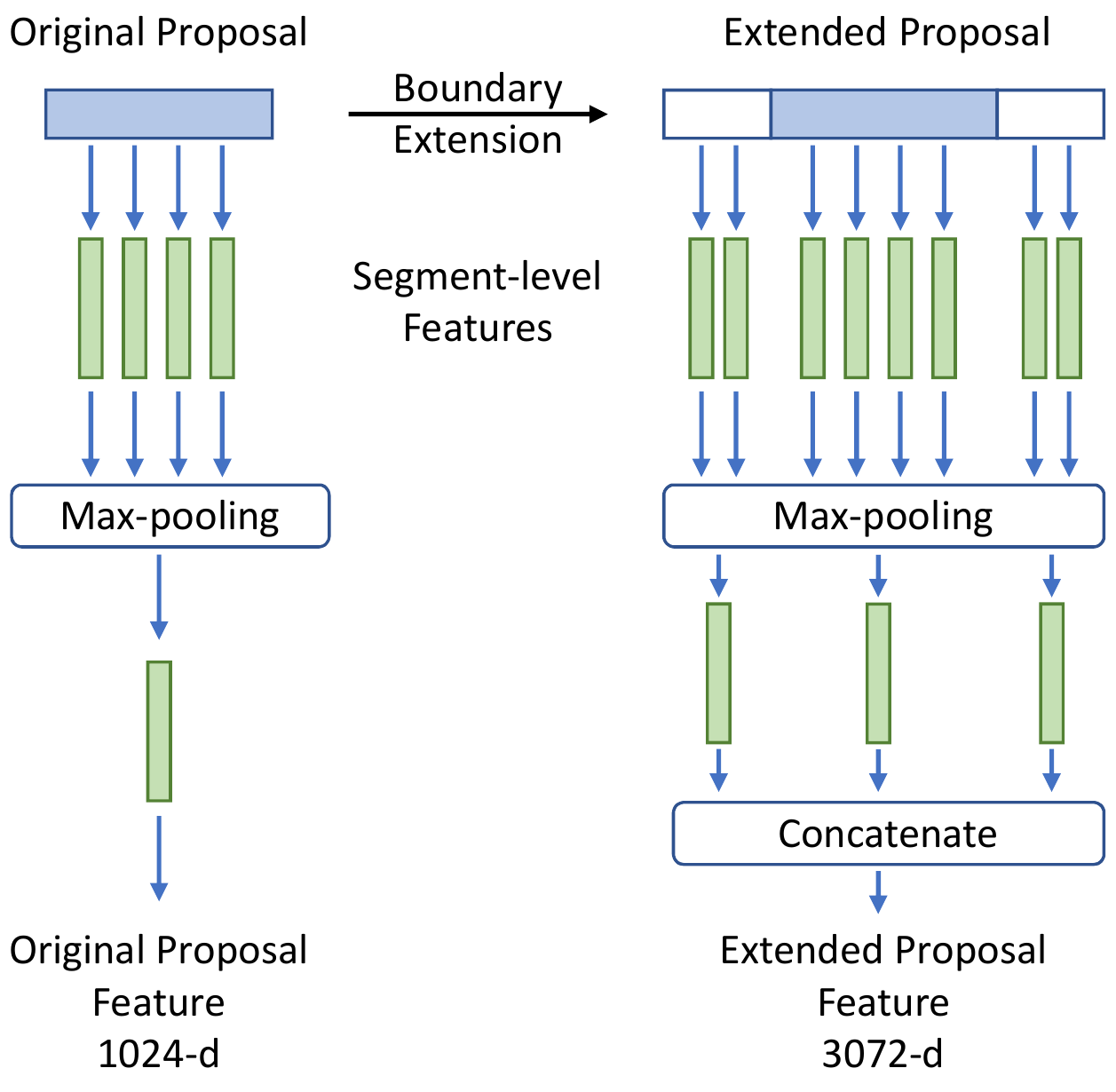}
	\caption{The illustration of (extended) proposal feature extraction.}
	\label{Fig:feature}
\end{figure}

We have two types of proposal features and the process of feature extraction is shown in Figure \ref{Fig:feature}.

\noindent \textbf{Proposal features.} For the original proposal, we first obtain a set of segment-level features within the proposal and then apply max-pooling across segments to obtain one 1024-dimensional feature vector.

\noindent \textbf{Extended proposal features.} The boundary of the original proposal is extended with $\frac{1}{2}$ of its length on both the left and right sides, resulting in the extended proposal. Thus, the extended proposal has three portions: \emph{start}, \emph{center} and \emph{end}. For each portion, we follow the same feature extraction process as the original proposal. Finally, the extended proposal feature is obtained by concatenating the feature of three portions.

\section{Network Architectures}

\begin{figure}[!tb]
	\centering
	\includegraphics[width=.85\linewidth]{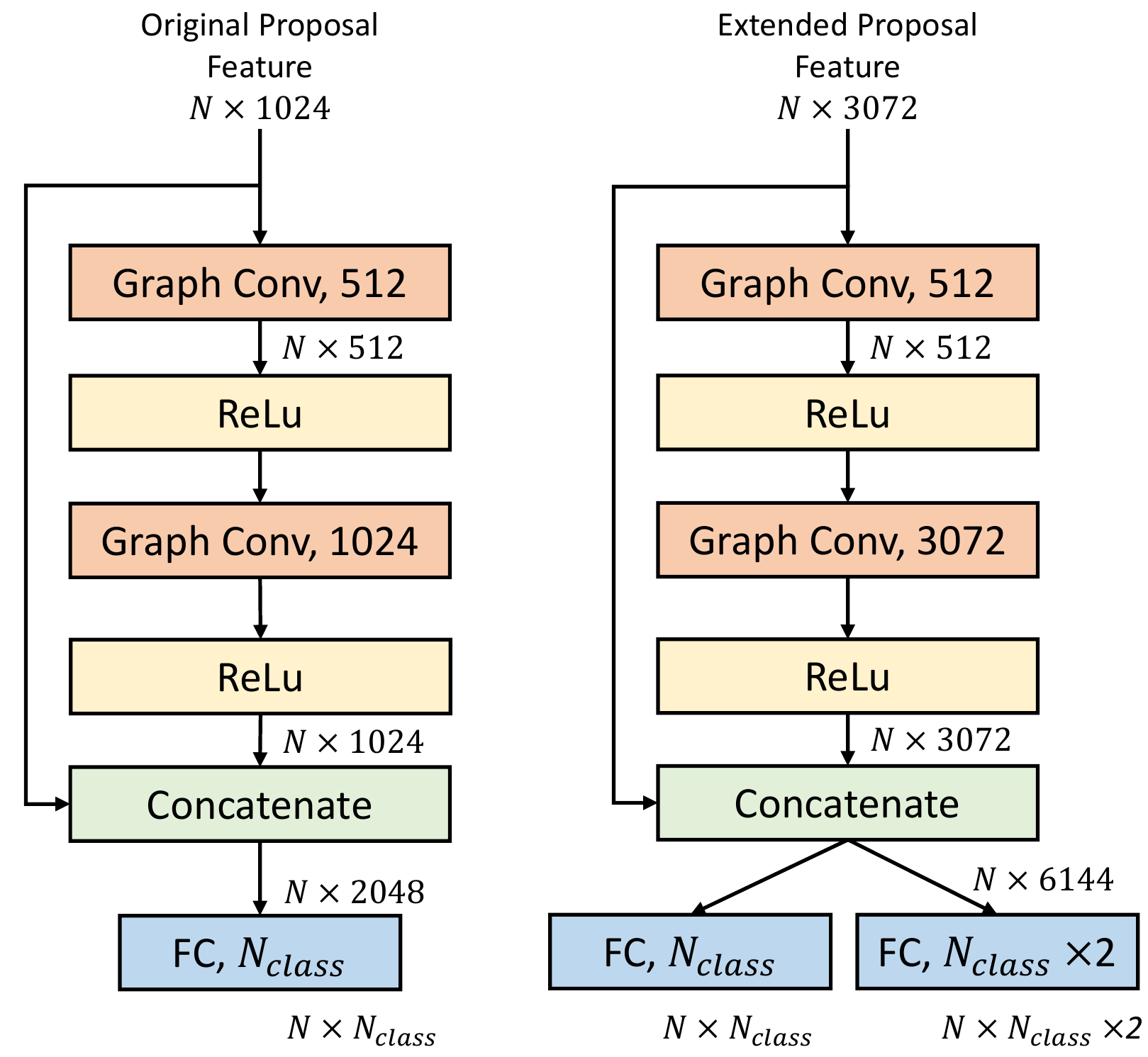}
	\caption{The network architecture of P-GCN model.}
	\label{Fig:network}
\end{figure}

\noindent \textbf{P-GCN.}
The network architecture of our P-GCN model is shown in Figure \ref{Fig:network}. Let $N$ and $N_{class}$ be the number of proposals in one video and the total number of action categories, respectively.
On the top of GCN, we have three fully-connected (FC) layers for different purposes. The one with $N_{class}\times 2$ outputs is for boundary regression and the other two with $N_{class}$ outputs are designed for action classification and completeness classification, respectively.

\begin{figure}[!bt]
	\centering
	\includegraphics[width=.9\linewidth]{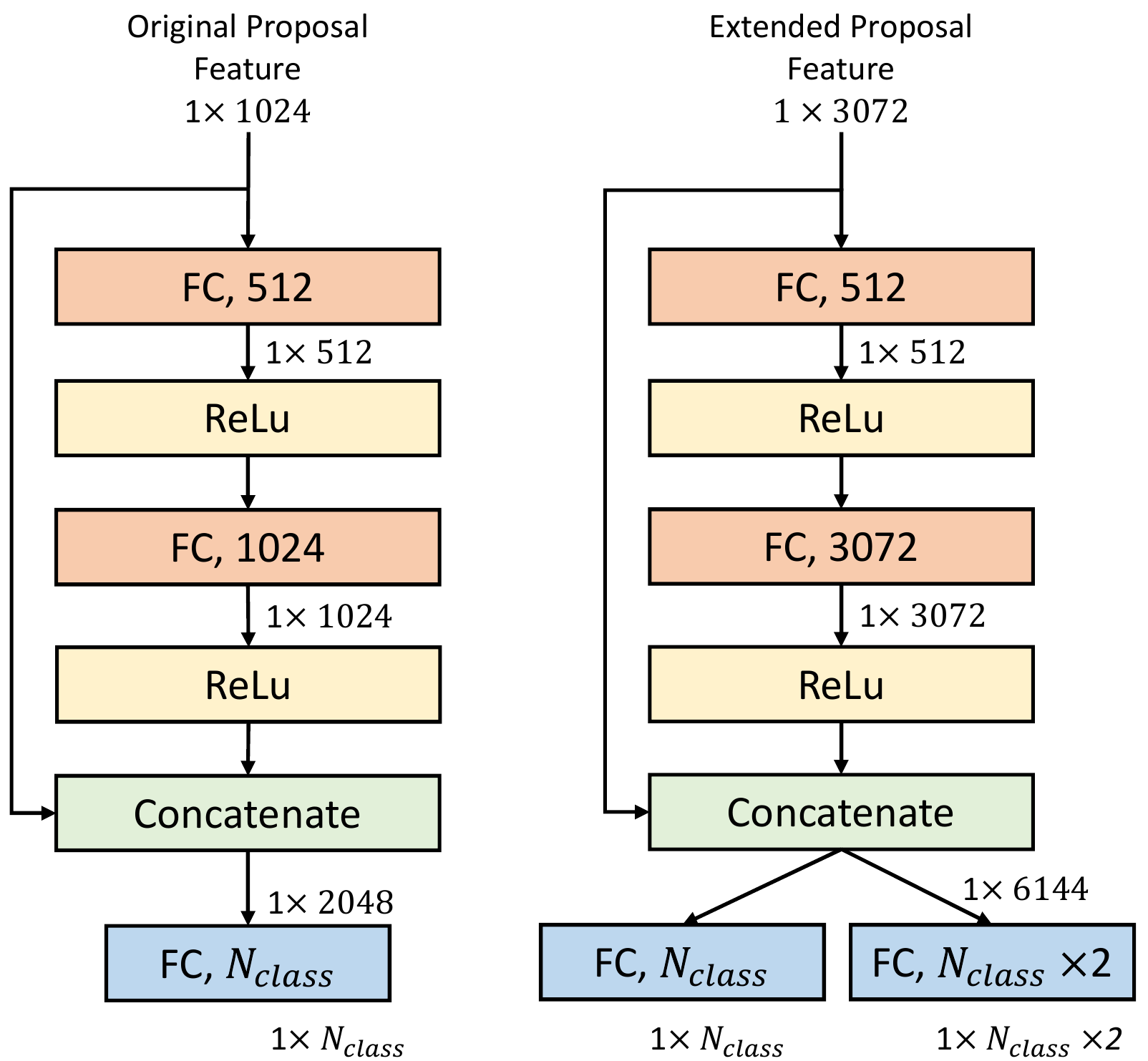}
	\caption{The network architecture of the MLP baseline.}
	\label{Fig:mlp_baseline}
\end{figure}

\begin{figure}[!bt]
	\centering
	\includegraphics[width=.9\linewidth]{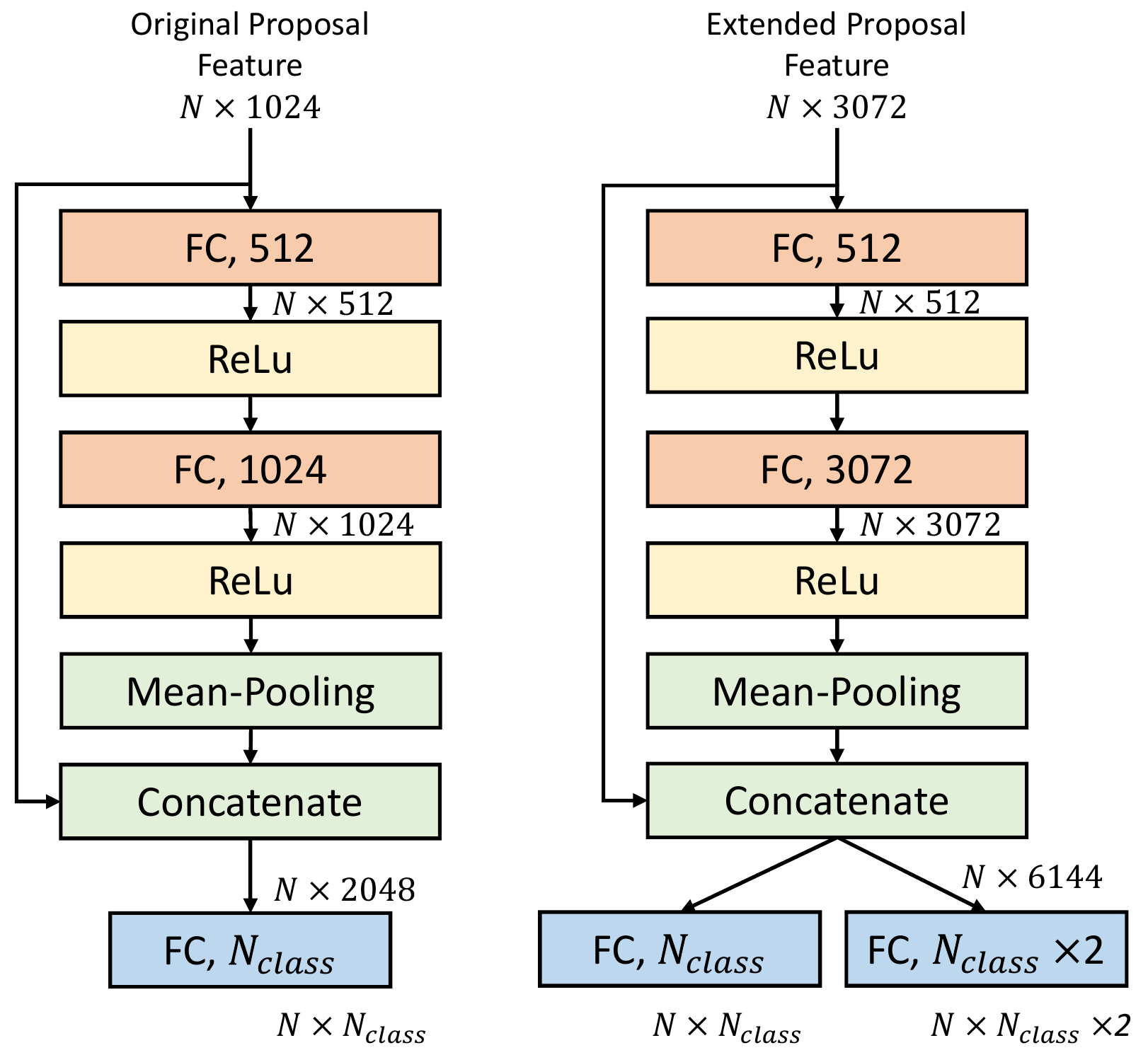}
	\caption{The network architecture of the Mean-Pooling baseline.}
	\label{Fig:mp_baseline}
\end{figure}


\noindent \textbf{MLP baseline.}
The network architecture of MLP baseline is shown in Figure \ref{Fig:mlp_baseline}.
We replace each of GCNs with a 2-layer multilayer perceptron (MLP). We set the number of parameters in MLP the same as GCN's for a fair comparison. Note that MLP processes each proposal independently without exploiting the relations between proposals. 


\noindent \textbf{Mean-Pooling baseline.}
As shown in Figure \ref{Fig:mp_baseline}, the network architecture of Mean-Pooling baseline is the same as the MLP baseline's except that we conduct mean-pooling on the output of MLP over the adjacent proposals.

\section{Training Details}

We have three types of training samples chosen by two criteria, \ie, the best tIoU and best overlap. For each proposal, we calculate its tIoU with all the ground truth in that video and choose the largest tIoU as the best tIoU (similarly for best overlap). For simplicity, we denote the best tIoU and best overlap as tIoU and OL. Then, three types of training samples can be described as: (1) Foreground sample: $tIoU\ge \theta_1$; (2) Incomplete sample: $OL\ge \theta_2,tIoU\le \theta_3$; (3) Background sample: $tIoU\le \theta_4$. These certain thresholds are slightly different on two datasets as shown in Table \ref{Tab:thr}. We consider all foreground proposals as the complete proposals.

\begin{table}[h]
	\centering
	\caption{The thresholds on different datasets.}
	\begin{tabular}{l|cccc}
		\hline
		Dataset                & $\theta_1$     & $\theta_2$  & $\theta_3$  & $\theta_4$        \\ \hline
		THUMOS14    & 0.7   & 0.7       & 0.3 & 0 \\
		ActivityNet v1.3   & 0.7   & 0.7      & 0.6 & 0.1 \\ \hline
	\end{tabular}
	\vspace{-0.3cm}
	\label{Tab:thr}
\end{table}

Each mini-batch contains examples sampled from a single video. The ratio of three types of samples is fixed to (1):(2):(3)=1:6:1. We set the mini-batch size to 32 on THUMOS14 and 64 on ActivityNet v1.3.

For more efficiency, we
fix the number of neighborhoods for each node to be 10 by
selecting contextual edges with the largest relevance scores
and surrounding edges with the smallest distances, where
the ratio of contextual and surrounding edges is set to 4:1.

In addition, we empirically found that setting $A_{i,j}$ to 0 (when  $A_{i,j}<0$) leads to better results.

\section{Loss function}
\noindent \textbf{Multi-task Loss.}
Our P-GCN model can not only predict action category but also refine the proposal’s temporal boundary by location regression. With the action classifier, completeness classifier and location regressors, we define a multi-task loss by:
\begin{equation}
\begin{aligned}
\mathcal{L}&=\sum_{i}\mathcal{L}_{cls}(y_i,\hat{y}_i)+\lambda_{1}\sum_{i}[y_i\ge1, e_i=1]\mathcal{L}_{reg}(o_i,\hat{o}_i) \\
&+\lambda_{2}\sum_{i}[y_i\ge1]\mathcal{L}_{com}(e_i,\hat{c}_i),
\end{aligned}
\end{equation}
where $\hat{y}_i$ and $y_i \in \{0,\dots,N_{class}\}$ is the predicted probability and ground truth action label of the $i$-th proposal in a mini-batch, respectively. Here, 0 represents the background class. $e_i$ is the completeness label.  $\hat{o}_i$ and $o_i$ are the predicted and ground truth offset, which will be detailed below. In all experiments, we set $\lambda_{1}=\lambda_{2}=0.5$.

\noindent \textbf{Completeness Loss.}
Here, the completeness term $\mathcal{L}_{com}$ is used only when $y_i\ge1$, \ie, the proposal is not considered as part of the background.

\noindent \textbf{Regression Loss.}
We devise a set of location regressors $\{R_m\}_{m=1}^{N_{class}}$, each for an activity category. For a proposal, we regress the boundary using the closest ground truth instance as the target. Our P-GCN model does not predict the start time and end time of each proposal directly. Instead, it predicts the offset $\hat{o}_i=(\hat{o}_{i,c}, \hat{o}_{i,l})$ relative to the proposal, where $\hat{o}_{i,c}$ and $\hat{o}_{i,l}$ are the offset of center coordinate and length, respectively. The ground truth offset is denoted as $o_i=(o_{i,c}, o_{i,l})$ and parameterized by:
\begin{equation}
\begin{aligned}
o_{i,c}&=(c_i-c_{gt})/l_i, \\
o_{i,l}&=log(l_i/l_{gt}), 
\end{aligned}
\end{equation}
where $c_i$ and $l_i$ denote the original center coordinate and length of the proposal, respectively. $c_{gt}$ and $l_{gt}$ account for the center coordinate and length of the closest ground truth, respectively.
$\mathcal{L}_{reg}$ is the smooth L1 loss and used when $y_i\ge1$ and $e_i=1$, \ie, the proposal is a foreground sample.

\section{Details of Augmentation Experiments on \\ActivityNet}

Our P-GCN model can be further augmented by taking the external video-level labels into account. To achieve this, we replace the predicted action classes in Eq. (6) with the external action labels. Specifically, given an input video, we use UntrimmedNet to predict the top-2 video-level classes and assign these classes to all the proposals in this video. In this way, each proposal has two action classes.

To further compute mAP, the score of each proposal is required. In our implementation, we follow the settings in BSN by calculating
\begin{equation}
s_{prop} = s_{act} * s_{com} * s_{bsn} * s_{unet},
\end{equation}
where $s_{act}$ and $s_{com}$ are the action score and completeness score associated with the action class.
$s_{bsn}$ represents the confidence score produced by BSN
and $s_{unet}$ denotes for the action score predicted by UntrimmedNet.

\section{Explanation and ablation study of $\theta_{ctx}$}
The parameter $\theta_{ctx}$ is a threshold value for constructing contextual edges, \ie $r(\Vec{p}_i, \Vec{p}_i)> \theta_{ctx}$. Since $r(\Vec{p}_i, \Vec{p}_i) \in [0,1]$, $ \theta_{ctx}$ can be chosen from $[0,1)$.   
An ablation study is shown in Table~\ref{tab:theta}. Our method performs well when $\theta_{ctx}=0.7,0.8,0.9$.

\begin{table}[!h]
	\footnotesize
	\centering
	\tabcolsep 4pt 
	\caption{\scriptsize{Results on THUMOS14 (Flow) with different $\theta_{ctx}$}.}
	\scalebox{0.75}
	{
		\begin{tabular}{c|cccccccccc}
			\hline
			$\theta_{ctx}$ & 0    & 0.1    & 0.2    & 0.3     & 0.4 & 0.5    & 0.6    & 0.7    & 0.8     & 0.9  \\ \hline
			mAP(tIoU=0.5) &
			45.31 & 45.29  & 45.37  & 45.61  & 45.65
			& 45.82  & 45.79  & 46.53  & 46.64   & 46.45 \\ \hline
		\end{tabular}
	}
	\label{tab:theta}
\end{table}




\section{Ablation study of boundary regression}
We conducted an ablation study on boundary regression in Table~\ref{tab:random}, whose results validate the necessity of using boundary regression.

\begin{table}[!h]
	\centering
	\caption{Ablation results of boundary regression on THUMOS14.}
	\begin{tabular}{l|c|c}
		\hline
		mAP@tIoU=0.5       & RGB    & Flow   \\ \hline
		without boundary regression & 36.4   & 45.4   \\
		with boundary regression      & \textbf{37.3}  & \textbf{46.5} \\ \hline
	\end{tabular}
	\label{tab:random}
	\vspace{-0.1in}
\end{table}

\begin{table}[!htb]
	\centering
	\caption{Runtime/computation complexity in FLOPs/action localization mAP on THUMOS14. We train each model with 200 iterations on a Titan X GPU and report the average processing time per video per iteration (note that proposal generation and feature extraction are excluded for each model). For P-GCN, we choose the number of sampling neighbourhoods as $N_s=4$.}
	\begin{tabular}{l|c|c|cc}
		\hline
		\multirow{2}{*}{Method} & \multirow{2}{*}{Runtime} & \multirow{2}{*}{FLOPs} & \multicolumn{2}{c}{mAP@tIoU=0.5} \\ \cline{4-5} 
		&                          &                        & RGB        & Flow       \\ \hline
		MLP baseline            & 0.376s                  & 16.57M                 & 34.8       & 43.7       \\
		P-GCN                   & 0.404s                  & 17.70M                 & 37.3       & 46.5       \\ \hline
	\end{tabular}
	\label{tab:runtime}
\end{table}

\section{Additional runtime compared to [52]}
The MLP baseline is indeed a particular implementation of [52], and it shares the same amount of parameters with our P-GCN. We compare the runtime between P-GCN and MLP in Table~\ref{tab:runtime}. It reads that GCN only incurs a relatively small additional runtime compared to MLP but is able to boost the performance significantly.

\end{document}